\documentclass[11pt]{article}

\usepackage[]{acl}

\usepackage{times}
\usepackage{latexsym}
\usepackage{multicol}
\usepackage{multirow}
\usepackage[T1]{fontenc}

\usepackage[utf8]{inputenc}

\usepackage{microtype}
\usepackage{amssymb}

\usepackage{inconsolata}

\usepackage{graphicx}
\usepackage[most]{tcolorbox}
\usepackage{listings}
\usepackage{booktabs}
\usepackage{multirow}
\usepackage{graphicx}

\usepackage{tabularx}
\usepackage{array}
\usepackage{enumitem}

\title{CoMMET: A Psychologically Grounded Benchmark for Evaluating Theory of Mind in Multimodal LLMs}

\author{\bf Ruirui Chen$^1$, Weifeng Jiang$^2$, Chengwei Qin$^{3}$, \\ \bf Kaiwen Wei$^{4}$, Yanzhen Yue$^{1}$, Cheston Tan$^{1}$\\
\textsuperscript{1}Institute of Advanced Intelligence and Computing (IAIC),\\
Agency for Science, Technology and Research (A*STAR), Singapore \\
\textsuperscript{2}Nanyang Technological University, Singapore \\
\textsuperscript{3}Hong Kong University of Science and Technology (Guangzhou), China \\
\textsuperscript{4}Chongqing University, China 
}

\begin{document}
\maketitle
\begin{abstract}
Theory of Mind (ToM)—the ability to reason about the mental states of oneself and others—is a cornerstone of human social intelligence. As Multimodal Large Language Models (MLLMs) become ubiquitous in real-world applications, validating their capacity for this level of social reasoning is essential for effective and natural interactions. However, existing benchmarks for assessing ToM in MLLMs are limited; most rely solely on text inputs and focus narrowly on belief-related tasks. In this paper, we propose a new multimodal benchmark dataset, CoMMET, a \textbf{Co}mprehensive \textbf{M}ental states and \textbf{M}oral \textbf{E}valuation \textbf{T}ask inspired by the Theory of Mind Booklet Task. CoMMET expands the scope of evaluation by covering a broader range of mental states and introducing multi-turn testing. To the best of our knowledge, this is the first psychology-grounded benchmark to evaluate MLLMs across multiple mental states in a multimodal, open-ended, and multi-turn setting. Through a comprehensive assessment of MLLMs across different families and sizes, we analyze the strengths and limitations of current models and identify directions for future improvement.\footnote{Dataset is available at: \url{https://huggingface.co/datasets/rrchen2026/Theory_of_Mind_CoMMET}}
\end{abstract}

\section{Introduction}

Theory of Mind (ToM) refers to the ability to attribute mental states to oneself and others \cite{Premack_Woodruff_1978, BARONCOHEN198537}. As Multimodal Large Language Models (MLLMs) have been widely adopted in daily life, validating whether MLLMs possess ToM capabilities, and to what extent, has become a popular research topic. Understanding how effectively MLLMs can interpret human mental states is essential for evaluating their reliability in applications aimed at improving human well-being.

\begin{figure*}[t]
    \centering
    \includegraphics[width=2\columnwidth]{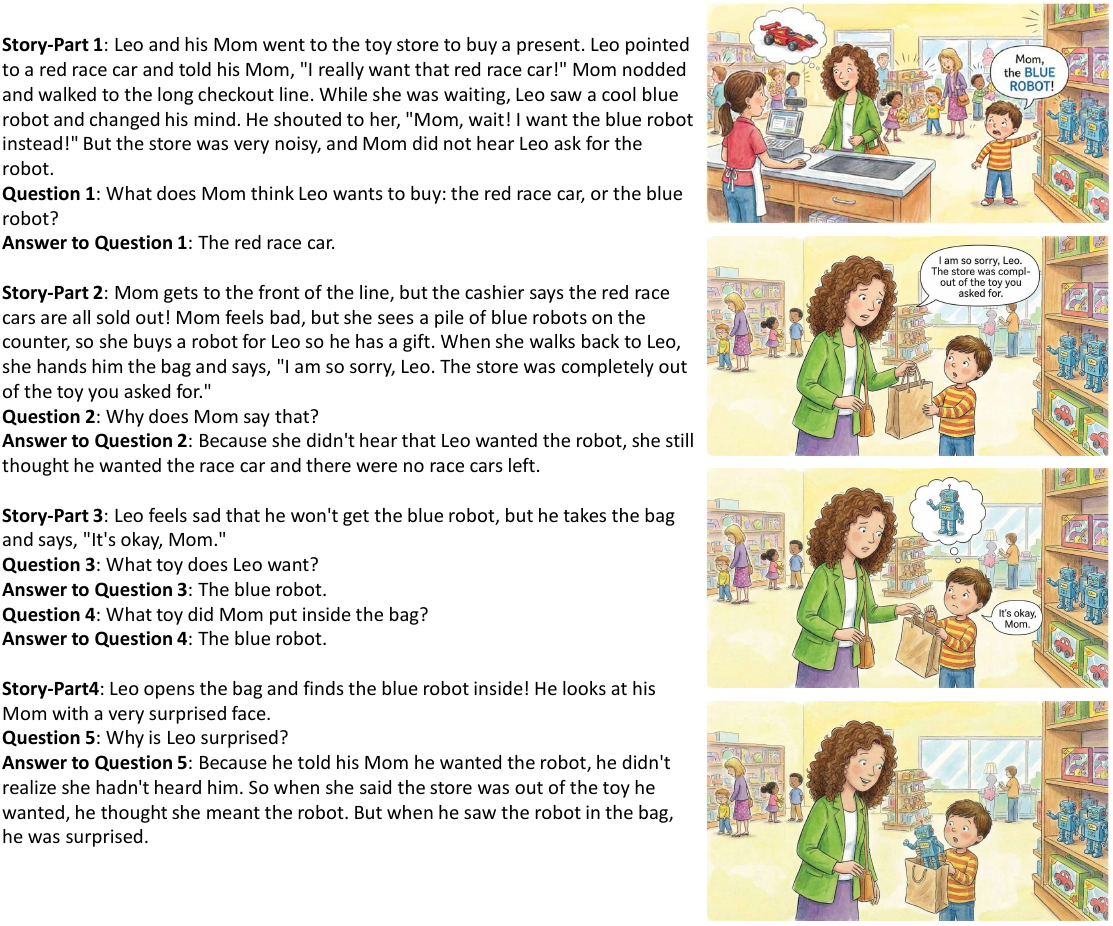}
    \caption{\textbf{An example of second-order belief–desire reasoning}. The story is presented across multiple rounds, with questions posed at different stages as the narrative unfolds. Images are provided for each round to enhance illustration and contextual understanding. In addition to the story content, we supply adaptive hints for MLLMs. For example, if the model answers the first question correctly, we add the feedback: "\textit{That’s right—Mom thinks Leo still wants the red race car.}" If the answer is incorrect, we provide a corrective hint: "\textit{Remember, Mom didn’t hear Leo say that he changed his mind, so she still believes he wants the first toy he picked.}"}
    \label{figure_1}
\end{figure*}
Theory of mind is a complex topic that encompasses multiple mental states \cite{beaudoin2020systematic}.
Numerous benchmarks have been proposed in recent years \cite{wu-etal-2023-hi, ma-etal-2023-tomchallenges, xu-etal-2024-opentom, chen-etal-2024-tombench, liu-etal-2025-tactfultom, villa-cueva-etal-2025-moments} to better understand the ToM capabilities of MLLMs. However, most existing benchmarks continue to focus primarily on belief-related reasoning, rely on text-only inputs \cite{chen-etal-2025-theory}, and adopt a single-round question–answering format \cite{wagner-etal-2025-mind, riemer2025position}. To address these limitations, we introduce a multimodal benchmark based on ToM booklet tasks \cite{Richardson2018DevelopmentOT, SOTOMAYORENRIQUEZ2024109905}, which covers a broader range of mental states and supports multi-round, interactive, story-based question answering.

ToM booklet tasks are established psychological assessments that cover a wide range of ToM concepts with varying levels of difficulty, from reasoning about desires and beliefs to reasoning about moral blameworthiness and mistaken referents. They also include items that assess the understanding of sarcasm and second-order belief–desire reasoning (Figure \ref{figure_1}). Building on this framework, our benchmark evaluates a broader spectrum of mental states and supports multi-turn evaluation within a story-based conversational setting.
To gain a deeper understanding of LLM performance on this task, we evaluate seven LLMs from four model families, including proprietary models as well as open-source models. We provide an in-depth analysis of model performance, highlighting their strengths and limitations, which helps identify directions for future improvement. Overall, our contributions are summarized as follows:
\begin{itemize}
    \item We introduce \textit{StoryTurn}, a multi-turn benchmark structure for facilitating LLM-based generation of multi-round data and evaluating LLMs in evolving scenarios.

    \item We present \textit{CoMMET}, a multimodal Theory of Mind benchmark covering seven ATOMS mental states \cite{beaudoin2020systematic} and moral reasoning, supporting both evaluation and capability improvement.
    
    \item Our analysis reveals distinct strengths and limitations of current LLMs across mental-state reasoning tasks, highlighting key challenges for future research.

\end{itemize}

\section{Related Work}
Theory of mind benchmarks have been broadly categorized into two types—story-based benchmarks and interactive benchmarks \cite{chen-etal-2025-theory}. As this paper primarily focuses on story-based benchmarks \cite{ma-etal-2023-towards-holistic}, we limit our review to benchmarks in this category.

In recent years, numerous story-based benchmarks have been proposed to evaluate the theory of mind capabilities of LLMs. Most existing benchmarks rely on text-only stories \cite{chen-etal-2024-tombench}, including narrative-style stories \cite{le-etal-2019-revisiting, wu-etal-2023-hi} and dialogue-based stories \cite{kim-etal-2023-fantom, chan-etal-2024-negotiationtom, liu-etal-2025-tactfultom}. These benchmarks typically focus on a narrow range of mental states, mostly focus on belief-related reasoning \cite{wu-etal-2023-hi} or reasoning about white lies \cite{liu-etal-2025-tactfultom}. A few multimodal benchmarks have also been proposed, such as MMToM-QA \cite{jin-etal-2024-mmtom} and MuMA-ToM \cite{Shi_Ye_Fang_Jin_Isik_Kuo_Shu_2025}; however, they mainly assess beliefs and intentions and are largely restricted to household settings.
MOMENTS \cite{villa-cueva-etal-2025-moments} is a multimodal video question-answering benchmark that covers all seven mental states defined in the ATOMS taxonomy. Nevertheless, it uses a multiple-choice QA format and evaluates LLMs in a single-turn setting without interaction.

A more detailed comparison with prior work covering multiple mental states is provided in Table \ref{compare_with_others} in Appendix \ref{comparison_with_existing}. Overall, existing benchmarks primarily rely on text-only inputs, cover a limited range of mental states, and evaluate models in single-turn, non-interactive settings.
\begin{figure}[t]
    \centering
    \includegraphics[width=\columnwidth]{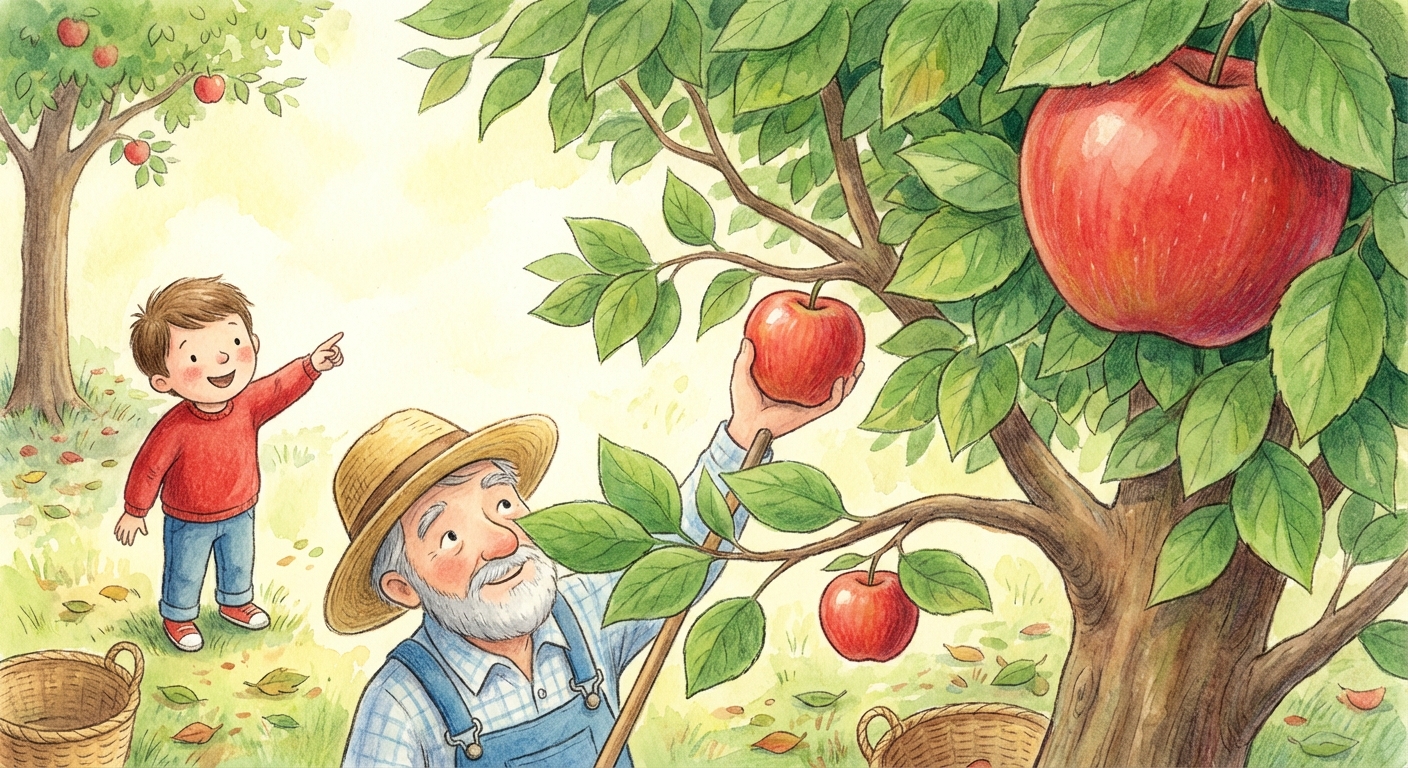}
    \caption{\textbf{A scenario analogous to a reference (hard) task from the ToM booklet task}. The child asks for the big apple; however, because the leaves block his view, Grandpa cannot see the largest apple at the top of the tree. As a result, he believes the medium apple is the biggest and therefore picks it for the child.}
    \label{figure_2}
\end{figure}

\section{CoMMET: Introduction and Design}
Building upon ToM booklet tasks \cite{Richardson2018DevelopmentOT, SOTOMAYORENRIQUEZ2024109905}, we develop a multimodal dataset to further evaluate the theory of mind capabilities of LLMs. In this section, we first provide a brief overview of ToM booklet tasks and then describe how our benchmark is constructed based on this established psychological framework.
\subsection{ToM Booklet Task}
Unlike the widely used Sally–Anne test \cite{BARONCOHEN198537} and Smarties task \cite{Perner1987ThreeyearoldsDW}, which primarily focus on belief reasoning, ToM booklet tasks are more challenging and assess a broader range of mental states. Although the ToM booklet task categorizes its test items differently from the ATOMS taxonomy \cite{beaudoin2020systematic}, they in fact cover all seven mental states defined by ATOMS. The booklet task includes both relatively simple items that test a single mental state and more complex items that require reasoning over multiple mental states simultaneously.

Figure~\ref{figure_2} illustrates a scenario similar to a reference (hard) task from the ToM booklet. In this scenario, there are three apples on a tree (small, medium, and big). However, Grandpa can see only two apples (small and medium); from his perspective, the medium apple is the largest. When the child asks for the big apple, Grandpa will pick the medium apple because he believes it is the biggest one available. This task begins with reasoning about Grandpa’s perceptual access (percepts), but answering the question "Which apple will Grandpa pick?" requires an additional step: modeling Grandpa’s internal mental representation and understanding that, from his belief, big refers to the medium apple.

In addition to assessing mental state reasoning, the task features advanced components like moral reasoning based on true and false beliefs (e.g., intent to harm). Furthermore, its multi-turn, conversational format (Figure~\ref{figure_1}) makes it more reflective of real-world scenarios. Given these strengths, the ToM booklet task provides a solid foundation for building a dataset that both measures and facilitates the further development of these capabilities.

\subsection{CoMMET}
Although the ToM booklet task is well suited for evaluating LLMs, it is limited in scale. To enable a more systematic assessment of LLMs’ theory of mind capabilities, we construct a dataset, CoMMET, based on the ToM booklet tasks.
\subsubsection{Multi-Turn Testing Format}
While conversational formats are natural for assessing children, evaluating LLMs requires a structured representation of multi-turn interactions, including the story context, feedback conditioned on previous responses (e.g., correctness-based feedback or alternative perspectives from other characters), current questions, images, and model responses. A structured format is also desirable for reliable LLM-based dataset generation. To support both multi-turn evaluation and effective dataset construction, we introduce the \textit{StoryTurn} format, in which each turn contains the following fields.
\begin{itemize}
    \item Story: The narrative text for this turn.
    \item Feedback: The feedback logic (e.g., "If right: ...; If wrong: ..." or "If the previous answer is A: ...; If the previous answer is B: ..."). Return empty string if not applicable.
    \item Question: The question to ask LLMs.
    \item Image: A unique Id for the illustration (e.g., "img01"). Return empty string if no image.
    \item Answer: The expected answer. Use semicolons to separate branching answers.
\end{itemize}
Below is an example of a StoryTurn that evaluates reasoning about "diverse desires". It consists of two turns, with the fields in each turn separated by "|". In this test, both answer sequences—(Apple, Chocolate) and (Chocolate, Apple)—are considered acceptable.
\begin{itemize}
    \item "Meet Sam. It is snack time! There are two snacks on the table. There is a shiny red apple, and there is a yummy chocolate bar. |  | which snack do you like best, the apple or the chocolate? | 1.png | Apple; Chocolate",
    \item "You do? That is great! But look, Sam likes the other snack better. |  |So which snack will Sam pick? The apple or the chocolate? |  | Chocolate; Apple"
\end{itemize}
\subsubsection{Dataset Construction}
\begin{figure*}[t]
    \centering
    \includegraphics[width=2\columnwidth]{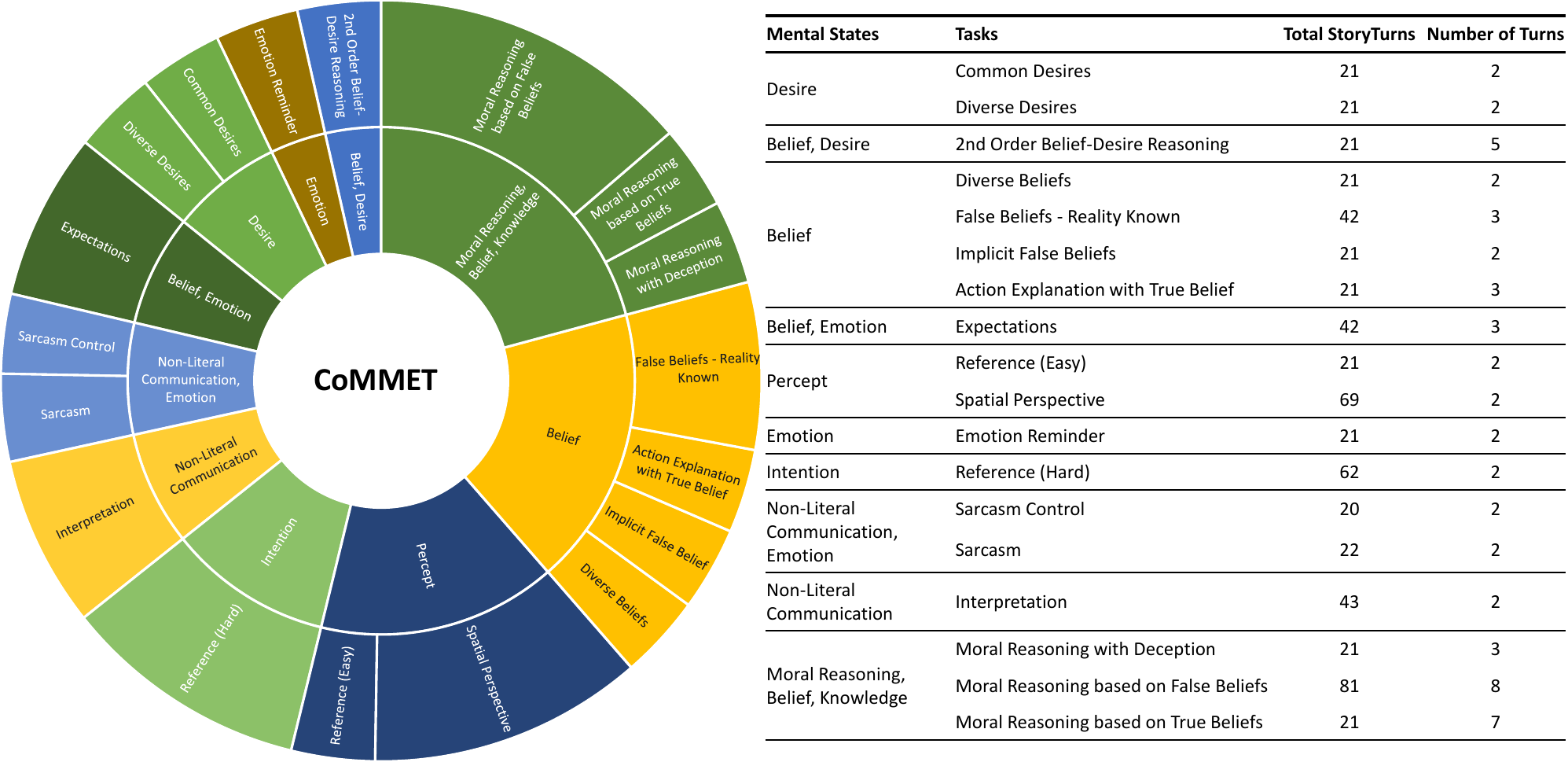}
    \caption{\textbf{Statistics of CoMMET}. To improve interpretability, we group related tasks from the Theory of Mind booklet and map them to the mental states defined in ATOMS. In addition to the mental states covered by ATOMS, the ToM booklet tasks also include moral reasoning. We report the number of StoryTurns for each task, as well as the number of turns within each StoryTurn.}
    \label{data_statistics}
\end{figure*}

To ensure both the diversity and accuracy of the dataset, we leverage a LLM to generate an initial dataset using one-shot learning. In this process, a single example StoryTurn from the ToM booklet task is provided to the LLM as a sample for generating new StoryTurns. After generation, we manually review all generated StoryTurn texts and images to validate their quality and correctness.
\paragraph{Dataset Generation}
As mentioned previously, rather than treating each question as a separate test item, we introduce StoryTurn to group related questions and evaluate LLMs in a story-based conversational setting. A total of 39 StoryTurns were constructed based on the original ToM booklet task.\footnote{\url{https://osf.io/g5zpv/}} Among them, 28 StoryTurns contain all the information required to answer the questions in the text alone, whereas the remaining 11 require visual information. In addition, we group the 39 StoryTurns into 18 tasks (Figure \ref{data_statistics}).

Through initial testing of the ToM booklet tasks on state-of-the-art LLMs, we found that Gemini 3.0 Pro\footnote{\url{https://aistudio.google.com/models/gemini-3}} achieved the best overall performance on both image-dependent and text-only StoryTurns, correctly answering 5 out of 8 image-based StoryTurns and 29 out of 31 text-based ones. In addition, Gemini 3.0 Pro Image, which is specifically designed to handle challenging image generation tasks, produces satisfactory images in most cases, as illustrated in Figure \ref{figure_1} and Figure \ref{figure_2}.
Therefore, our pipeline first generates StoryTurns for each task in the original ToM booklet dataset using one-shot prompting with Gemini 3.0 Pro, and then produces corresponding illustrative images for each StoryTurn based on selected segments of the story. The detailed prompts are shown below.
\begin{tcolorbox}[
    colback=gray!5,
    colframe=gray!55,
    coltitle=black,
    colbacktitle=gray!55,
    title=Prompt for Generating Text Stories in StoryTurn Format,
    fonttitle=\bfseries\small,
    fontupper=\ttfamily\scriptsize,
    boxsep=0mm,
    left=1.5mm,
    right=1.5mm,
    top=1mm,
    bottom=1mm,
    toptitle=0.5mm,
    bottomtitle=0.5mm,
    arc=1.5mm,
    boxrule=0.4pt,
]
You are an expert at creating logic puzzles for children (Grade 2).

Task: Create a NEW, DISTINCT multi-turn story testing "\{aspect\}".
        
Format Reference:
\{original\_content\}
        
Requirements:

1. Create a completely different scenario.

2. STRUCTURE: The story must have multiple turns. 

3. IMAGES: Mirror the image structure of the original example exactly. If a turn has a single image, generate one for the new story. If a turn contains two images separated by a semicolon (;), you must generate two corresponding images to represent the different conditions.

4. IDs: Use unique Image IDs starting with "\{id\_prefix\}\_".
\end{tcolorbox}

\begin{tcolorbox}[
    colback=gray!5,
    colframe=gray!55,
    coltitle=black,
    colbacktitle=gray!55,
    title=Prompt for Image Generation,
    fonttitle=\bfseries\small,
    fontupper=\ttfamily\scriptsize,
    boxsep=0mm,
    left=1.5mm,
    right=1.5mm,
    top=1mm,
    bottom=1mm,
    toptitle=0.5mm,
    bottomtitle=0.5mm,
    arc=1.5mm,
    boxrule=0.4pt,
]
Children's book illustration.

PREVIOUS STORY EVENTS (Context): \{story\_history\}

CURRENT SCENE: \{specific\_context\}

CONTINUITY: The illustration should naturally follow the previous image as the next sequential scene, while keeping the characters' appearances exactly consistent.
\end{tcolorbox}
\paragraph{Dataset Validation}
We manually reviewed all generated StoryTurn texts. Overall, the texts were largely consistent with the intended task categories, with only one misclassified StoryTurn: it was assigned to the non-sarcasm task but actually belonged to the sarcasm task. The main textual issue was that the model occasionally included feedback in the same turn as the answer, rather than deferring it to the subsequent turn as required by the multi-turn setting. In addition, the model sometimes made the story context overly explicit by revealing information related to the correct answer, thereby reducing task difficulty.

A more prominent challenge arose during the image-generation process. Detailed examples of image-related issues are provided in Appendix~\ref{issues_image_generation}. To ensure the quality of the dataset, one human expert jointly reviewed all textual and visual content according to the criteria listed in Table~\ref{validation_criteria}. The expert manually corrected the identified issues in the StoryTurn text and regenerated problematic images using Gemini Pro through the web interface\footnote{\url{https://gemini.google.com/app}} to improve consistency between each image and its corresponding story. Subsequently, two additional human experts independently reviewed each StoryTurn, including both its textual and visual content, to ensure that it was error-free and internally consistent. Textual issues were corrected directly, whereas problematic images were regenerated iteratively until all three experts agreed that they were free of errors and aligned with the corresponding text. In total, approximately 567 of the 844 images were manually revised. The detailed distribution of the identified errors is presented in Table~\ref{image_error_statistics} in Appendix \ref{issues_image_generation}.

\subsubsection{Dataset Statistics}
After filtering process, we obtained 591 StoryTurns, comprising 1,976 questions and 844 images, to evaluate LLMs across a diverse range of mental states.
The average StoryTurn in CoMMET comprises 3.34 conversational turns with questions, with some narratives extending to as many as eight turns. Notably, most StoryTurns probe multiple mental states. For clarity of analysis, each StoryTurn is categorized according to its dominant mental state. As illustrated in Figure~\ref{data_statistics}, CoMMET spans a broad spectrum of tasks, ranging from simple to complex, and covers all mental states defined in ATOMS, in addition to moral reasoning tasks grounded in false beliefs or intent to cause harm.

Furthermore, as shown in Figure~\ref{data_statistics} and Table \ref{compare_with_others} in Appendix \ref{comparison_with_existing}, CoMMET incorporates tasks that probe contradictory reasoning (e.g., common versus divergent desires), includes explicit "why" questions, and features StoryTurns that simultaneously evaluate multiple mental states, such as second-order belief–desire reasoning. Together, these characteristics enable a more precise assessment of whether LLMs genuinely comprehend the underlying tasks.

\begin{table*}[t]
  \centering
    \renewcommand\arraystretch{1}
  \setlength{\tabcolsep}{1.5mm}{
  \resizebox{\textwidth}{!}{
    \begin{tabular}{lccccccccccc}
    \toprule
    \textbf{Models} & \textbf{Image} & \textbf{Overall} & \textbf{Belief} & \textbf{Emotion} & \textbf{Intention} & \textbf{Desire} & \textbf{Percept} & \textbf{Belief\_Emotion} & \textbf{Belief\_Desire} & \textbf{NLCE}  & \textbf{MBK} \\
    \midrule
    \multirow{2}[0]{*}{Human} & No & 76.07 & 84.38 & 75.00& 66.67 & 100.00 & 87.50 & 62.50 & 25.00 & 75.00 & 66.67  \\
          & Yes & 80.99& 84.38 & 87.50 & 66.67 & 100.00 &  87.50& 75.00 & 25.00 & 87.50 & 77.27  \\
    \midrule
    \multirow{2}[0]{*}{LLaMA-3.2-11B} & No & 44.86 & 55.24 & 80.95 & 33.33 & 97.62 & 42.86 & 59.52 & 14.29 & 38.10 & 3.57  \\
          & Yes & 47.12 & 63.81 & 90.48 & 47.62 & 92.86 & 28.57 & 52.38 & 4.76 & 50.00 & 3.57  \\
    \midrule
    \multirow{2}[0]{*}{LLaMA-4-Maverick} & No & 55.89 & 74.29 & 85.71 & 38.10 & 100.00 & 42.86 & 88.10 & 33.33 & 52.38 & 2.38  \\
          & Yes & 55.89 & 67.62 & 80.95 & 23.81 & 100.00 & 42.86 & 83.33 & 38.10 & 61.90 & 11.90  \\
    \midrule
    \multirow{2}[0]{*}{Claude-Haiku-4.5} & No & 76.94 & 86.67& 100.00 & 52.38 & 92.86 & 90.48 & 97.62 & 4.76 & 78.57 & 60.71  \\
          & Yes & 80.70 & 94.29 & 100.00 & 42.86 & 100.00 & 100.00 & 95.24 & 14.29 & 88.10 & 59.52 \\
    \midrule
    \multirow{2}[0]{*}{Claude-Opus-4.7} & No & 93.23 & 98.10 & 100.00 & 90.48 & 100.00 & 85.71 & 100.00 & 57.14 & 92.86 & 90.48  \\
          & Yes & 92.48 & 98.10 & 100.00 & 95.24 & 100.00 & 90.48 & 100.00 & 28.57 & 95.24 & 90.48  \\
    \midrule
    \multirow{2}[0]{*}{GPT-4o} & No & 78.70 & 92.38 & 100.00 & 66.67 & 100.00 & 85.71 & 92.86 & 28.57 & 85.71 & 48.81   \\
          & Yes & 82.46 & 98.10 & 100.00 & 61.90 & 100.00 & 80.95 & 95.24 & 28.57 & 88.10 & 59.52   \\
    \midrule
    \multirow{2}[0]{*}{GPT-5.4} & No & 87.47 & 96.19 & 100.00 & 76.19 & 100.00 & 100.00 & 95.24 & 47.62 & 95.24 & 69.05   \\
          & Yes & 87.97 & 96.19 & 100.00 & 85.71 & 100.00 & 95.24 & 95.24 & 57.14 & 95.24 & 67.86  \\
    \midrule
    \multirow{2}[0]{*}{Gemini-3.1-Flash-Lite} & No & 84.71 & 93.33 & 95.24 & 42.86 & 100.00 & 80.95 & 95.24 & 38.10 & 95.24 & 76.19 \\
          & Yes & 86.47 & 94.29 & 100.00 & 61.90 & 100.00 & 90.48 & 90.48 & 42.86 & 90.48 & 78.57 \\
    \bottomrule
    \end{tabular}%
    }}
    \caption{\textbf{Story-level accuracy of different LLMs on the CoMMET subset containing complete textual information}. Two experimental settings are considered: (1) text-only input and (2) text with illustrative images. Results are reported for overall performance as well as performance across different mental-state categories. NLC denotes \textit{Non-Literal Communication}, NLCE denotes \textit{Non-Literal Communication with Emotion}, and MBK denotes \textit{Moral Reasoning, Belief, and Knowledge}. The values are reported as percentages.
}
  \label{text_complete}%
\end{table*}%
\section{Experiments}

In this section, we describe the experimental setup and evaluation methodology, including the multimodal large language models evaluated and the dataset used. We then provide a detailed analysis of the experimental results, examining model performance across different task categories, mental states, and levels of complexity, and discuss the broader implications of these findings.
\subsection{Experimental Setup}
\paragraph{Models}
Given the challenging nature of some tasks in CoMMET, we mainly focus on state-of-the-art models that support multimodal inputs, covering a diverse range of families, including proprietary and open-source models, as well as thinking models and standard models. Specifically, we evaluate models from four major model families: GPT-5.4 and GPT-4o\footnote{\url{https://openai.com}} from OpenAI; Gemini-3.1-Flash-Lite\footnote{\url{https://ai.google.dev/gemini-api}} from Google; Claude-Haiku-4.5 and Claude-Opus-4.7\footnote{\url{https://www.anthropic.com/claude}} from Anthropic; and LLaMA-4-Maverick and LLaMA-3.2-11B Vision-Instruct\footnote{\url{https://developer.meta.com/ai/models}} from Meta. We include LLaMA-3.2-11B Vision-Instruct to examine how a smaller-scale vision-language model performs on these tasks.

\paragraph{Dataset}
As mentioned previously, CoMMET contains a total of 591 StoryTurns. Among these, 399 StoryTurns provide all the necessary information in text form (text-complete subset), while the remaining 192 require complementary information from images to answer specific questions (image-dependent subset). Consequently, we conduct experiments and perform analyses separately on these two subsets of StoryTurns.

\paragraph{Evaluation}
In this paper, we report story accuracy \cite{le-etal-2019-revisiting} at multiple levels of granularity, including overall performance across all tasks and accuracy within each mental state category. Story accuracy is defined such that a story is considered correctly answered only if all questions associated with that story are answered correctly. 

Given that our dataset consists of open-ended question-answering tasks and requires large-scale evaluation across diverse task types and LLMs, we adopt an LLM-based evaluation strategy. Based on the high agreement between human experts and Gemini 3.1 Pro (Appendix \ref{justification_of_llm_evaluator}), we use Gemini 3.1 Pro as the judge model to compare each model response with the corresponding ground-truth answer and determine its semantic correctness.

\subsection{Main Results}
Tables \ref{text_complete} and \ref{text_incomplete} present the performance of different LLMs on two CoMMET subsets: the subset in which all information required to answer the questions is contained in the text, and the subset in which visual information is required. From the results, we draw the following observations:
\begin{itemize}
\item Achieving high story-level accuracy remains challenging, as few LLMs attain 90\% story-level accuracy on either subset.
\item CoMMET tasks are particularly difficult for smaller models (e.g., LLaMA-3.2-11B-Vision-Instruct), indicating that further improvements are needed to enhance their performance on this type of reasoning task.
\item When all necessary information is already present in the text, adding illustrative images still leads to overall performance improvements for most models, particularly in tasks involving non-literal communication and emotion understanding, as shown in Table \ref{text_complete}.
\item Most LLMs perform well on simpler emotion and desire tasks, achieving near-perfect accuracy. However, further improvements are needed for more complex tasks involving multiple mental states.
\end{itemize}

\begin{table}[t]
  \centering
  \renewcommand\arraystretch{1.1}
  \setlength{\tabcolsep}{1.5mm}{
  \resizebox{\columnwidth}{!}{
    \begin{tabular}{lccccc}
  \toprule
    \textbf{Models} & \textbf{Overall} & \textbf{Percept} & \textbf{Intention} & \textbf{MBK} & \textbf{NLC}\\
    \midrule
    Human & 52.00 & 68.75&  40.00& 25.00 &  56.25\\
    \midrule
    LLaMA-3.2-11B & 17.19 & 18.84 & 9.76 & 0.00 & 37.21 \\
    LLaMA-4-Maverick & 35.42 & 40.58 & 9.76 & 41.03 & 46.51 \\
    Claude-Haiku-4.5 & 45.83 & 43.48 & 19.51 & 53.85 & 67.44 \\
    Claude-Opus-4.7 & 62.50 & 60.87 & 26.83 & 71.79 & 90.70 \\
    GPT-4o & 48.44 & 49.28 & 9.76 & 61.54 & 72.09 \\
    GPT-5.4 & 53.65 & 53.62 & 24.39 & 66.67 & 69.77 \\
    Gemini-3.1-Flash-Lite & 57.29 & 53.62 & 29.27 & 71.79 & 76.74 \\
    \bottomrule
    \end{tabular}}%
    }
    \caption{\textbf{Story-level accuracy of different LLMs on the CoMMET subset where visual information is required to answer the questions}. We report both overall performance and performance across individual mental-state categories. MBK denotes \textit{Moral Reasoning, Belief, and Knowledge}. The values are reported as percentages}
  \label{text_incomplete}%
\end{table}%
\begin{table}[t]
\centering
\renewcommand{\arraystretch}{1.05}
\setlength{\tabcolsep}{1.5mm}{
\resizebox{\columnwidth}{!}{
\begin{tabular}{lccc}
\toprule
\multirow{2}{*}{\textbf{Models}}
& \multicolumn{2}{c}{\textbf{VI Not Required}}
& \multicolumn{1}{c}{\textbf{VI Required}} \\
\cmidrule(lr){2-3}
\cmidrule(lr){4-4}
& With Image & No Image & With Image \\
\midrule
Human                    & 3/3   & 1/1   & 2/6  \\
\midrule
LLaMA-3.2-11B            & 14/24 & 22/29 & 22/40 \\
LLaMA-4-Maverick         & 8/14  & 13/19 & 9/32  \\
Claude-Haiku-4.5         & 11/12 & 13/14 & 14/30 \\
Claude-Opus-4.7          & 1/1   & 2/2   & 17/27 \\
GPT-4o                   & 6/7   & 7/7   & 12/37 \\
GPT-5.4                  & 3/3   & 5/5   & 13/27 \\
Gemini-3.1-Flash-Lite    & 6/8   & 10/11 & 8/23  \\
\bottomrule
\end{tabular}}
}
\caption{Effect of feedback in correcting LLM response trajectories. We report the proportion of cases in which an LLM answered a question incorrectly but answered the subsequent question correctly after receiving feedback, among all cases where an incorrect answer was followed by feedback. VI denotes \textit{visual information}.}
\label{feedback}
\end{table}

\subsection{Effectiveness of Feedback}

We examine whether correctness-based feedback helps LLMs recover from incorrect reasoning in multi-turn theory of mind tasks. Specifically, we consider cases where a model answers a question incorrectly and then receives feedback before the next turn. Feedback is considered effective if the model answers the subsequent question correctly. Since different LLMs achieve different overall performance and therefore produce different numbers of incorrect answers, Table~\ref{feedback} reports the proportion of effective feedback cases among all incorrectly answered turns followed by correctness-based feedback. This metric indicates how often correctness-based feedback redirects models from an incorrect response trajectory to a correct one. As shown in the table, most LLMs can use feedback to revise their subsequent answers, although the degree of improvement varies across models.
\subsection{Do LLMs Truly Understand the Tasks?}
While LLMs demonstrate near-perfect performance on relatively simple tasks, their effectiveness declines significantly on more challenging tasks, especially those requiring explanatory ("why") reasoning.

\paragraph{Belief and Desire}
Taking the second-order belief–desire task (Figure \ref{figure_1}) as an example, we observe that after validation, most errors occur in the final turn, when the model is asked why a character is surprised or confused. The most common error is answering that he believed his mother did not know he had changed his mind. Notably, even Gemini 3.1 Flash and Gemini 3.1 Pro exhibit this error pattern, calling into question whether LLMs genuinely possess ToM capabilities.
\begin{figure}[t]
    \centering
    \includegraphics[width=\columnwidth]{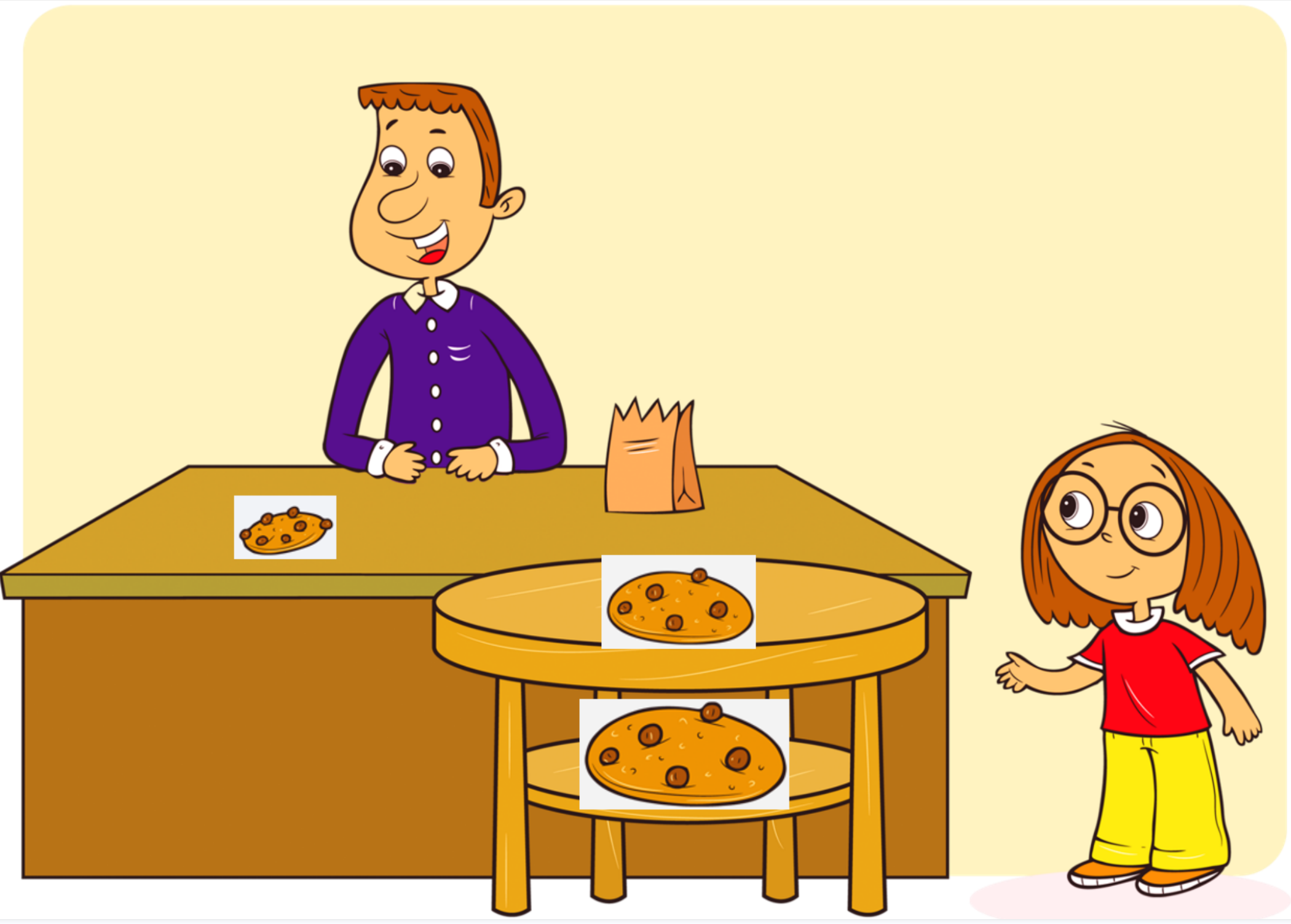}
    \caption{\textbf{A scenario from the Reference (Hard) task in the ToM Booklet} \cite{Richardson2018DevelopmentOT, SOTOMAYORENRIQUEZ2024109905}. Although the child asks for the big cookie, the father—due to his viewing position—can see only the small and medium cookies. Consequently, he believes the medium cookie is the largest available and therefore packs it for the child.}
    \label{figure_4}
\end{figure}
\paragraph{Intention and Perspective-Taking}
We take the hard reference task—an image-dependent task from the ToM Booklet—as an example. In this scenario, illustrated in Figure \ref{figure_4}, Marie’s mother leaves three cookies of different sizes in the kitchen: a small cookie on the counter, a medium-sized cookie on the table, and a large cookie on a shelf. Marie and her father then enter so that her father can pack her a snack for school. Marie tells her father, "I want the big cookie." The question is: "Which cookie will the father pack for Marie?"
Most models select the large cookie, including Gemini 3.1 Pro and GPT-5.4. When we subsequently provide the information "Actually, the father packs the medium cookie" and ask "Why?", these models respond with explanations such as "The big one may be too large for a snack, so the medium one is more appropriate."
LLaMA-4-Maverick produces the correct answer to the first question; however, its explanation for the second question—"Because Marie’s mom said to take the one in the middle"—is entirely hallucinated.

\paragraph{Belief}
Unlike the widely studied false-belief task in the Sally–Anne test and Smarties task, the ToM Booklet includes a diverse belief task. In this task, a child searches for an object that may be located in either location A or location B. The first question asks LLMs to predict where the object actually is. Next, the model is told that the child believes the object is not at that location, and the second question asks where the child will go to look for the object. In the third turn, the child fails to find the object at the location he/she believed it to be, and the model is asked to infer the object’s true location.
Our evaluation shows that LLMs frequently make errors in this third turn. Models sometimes predict a location other than A or B, or respond that the location is unknown. This pattern suggests that LLMs do not robustly distinguish between reality and belief in this task.

\paragraph{Summary} In conclusion, although LLMs demonstrate strong performance on certain relatively simple tasks, we argue that they do not yet exhibit consistent or robust theory of mind capabilities. Their success is often task-specific and does not reliably generalize to more complex scenarios involving diverse beliefs, perspective-taking, or explanatory reasoning. This observation aligns with findings from prior studies \cite{ ullman2023large, kim-etal-2023-fantom, 10.1145/3610978.3640767, 10.5555/3692070.3694663, bortoletto-etal-2024-limits, shapira-etal-2024-clever, chen-etal-2024-tombench}., which similarly report that apparent competence in isolated tasks does not necessarily reflect a genuine or stable theory of mind understanding.
\subsection{Human Study}
To address concerns about the interpretability of CoMMET scores, we collected human responses under the same task format and scoring protocol as shown in Table \ref{text_complete}, Table \ref{text_incomplete} and Table \ref{feedback}. The details about how we test humans are shown in Appendix \ref{human_study}.  Unlike multi-choice question answering tasks, CoMMET uses open-ended question answering, which is more challenging for participants because they must formulate complete responses rather than select from predefined options. We use these results as a human reference baseline rather than as a strict upper bound. Interestingly, human participants do not always outperform LLMs, particularly on several challenging multi-turn tasks. The main reason is that adult participants may overthink the given story or overlook specific details while reading it. Two illustrative examples are provided in Appendix \ref{human_study}. This pattern is consistent with recent findings that frontier LLMs can match or exceed sampled human participants on some text-based or highly structured ToM tasks \cite{strachan2024testing}, although they may still fail on more interactive, coherent, or action-oriented ToM settings. 
Together with the dataset’s derivation from established psychological ToM paradigms, manual validation, and task-category design, the human baseline provides additional evidence that CoMMET measures non-trivial social reasoning abilities rather than simple language understanding alone.
\section{Conclusion}
In this paper, we propose CoMMET, a benchmark inspired by the psychological theory of mind booklet tasks. CoMMET is a multimodal dataset that encompasses all the mental states defined in ATOMS and additionally includes moral reasoning. Unlike traditional approaches that present the complete story and ask questions only at the end, CoMMET evaluates MLLMs across multiple conversational turns, providing feedback based on their answers to previous questions. This approach offers a more naturalistic evaluation, akin to how children are tested in real-world scenarios.
To enhance interpretability and reduce ambiguity, we map the diverse tasks in CoMMET to their corresponding ATOMS mental states. We evaluate a broad range of MLLMs from multiple model families, including both state-of-the-art and smaller models, to examine how model scale influences performance across these tasks. Through detailed comparisons and analyses, we identify key areas requiring further improvement. We believe that our benchmark dataset, together with the limitations revealed in this study, offers valuable insights and concrete guidance for future research on advancing robust theory of mind reasoning in LLMs.

\section*{Limitations}
Although we select GPT-3.1-Pro as the evaluator due to its highest agreement with human judgments, it remains imperfect. Developing more accurate and reliable automated evaluation methods for open-ended explanations is therefore an important direction for future work. In addition, CoMMET remains a story-based benchmark. Extending it to interactive settings and incorporating richer modalities, such as acoustic information, would enable more realistic evaluation of MLLMs' theory of mind capabilities and constitutes another promising direction for future research.

\section*{Ethical Considerations}

This work introduces and evaluates CoMMET, a benchmark for assessing theory of mind capabilities in multimodal large language models. The benchmark is designed for research purposes and should not be interpreted as evidence that current models possess human-like mental states, consciousness, or genuine social understanding. Instead, model performance reflects their ability to generate answers that align with expected interpretations of mental-state reasoning tasks.

The benchmark includes scenarios involving beliefs, emotions, intentions, desires, perceptions, knowledge, non-literal communication, and moral reasoning. These topics may involve socially sensitive situations or interpersonal judgments. During data construction and validation, we aimed to avoid content that could promote harmful stereotypes, discriminatory assumptions, or inappropriate judgments about individuals or groups. We also manually checked the generated stories and images to reduce ambiguity, factual inconsistency, and potentially misleading visual information.

For human evaluation, participants were recruited voluntarily and anonymously. We did not collect personally identifiable information such as names, identification numbers, IP addresses, device information, or browser metadata. Participants were informed that the task was for research purposes and could stop participation at any time. The collected responses were used only to estimate human performance on a subset of CoMMET and to provide a baseline for model evaluation.

We acknowledge that ToM-related benchmarks may be misused to claim that models can reliably infer human beliefs, intentions, or emotions in real-world settings. Such claims would be inappropriate, especially in high-stakes domains such as healthcare, education, hiring, legal decision-making, or mental health support. Performance on CoMMET should therefore be understood as task-specific behavioral evidence rather than a guarantee of robust social reasoning in open-ended real-world interactions. We encourage future users of the benchmark to report limitations clearly and to avoid deploying models for sensitive interpersonal inference without appropriate human oversight.

\bibliography{custom}

\appendix
\clearpage
\newpage
\section{Comparison with Existing Benchmarks} \label{comparison_with_existing}
Table~\ref{compare_with_others} compares CoMMET with five recently published benchmarks that cover more than one mental states defined in ATOMS, including ToMBench \cite{chen-etal-2024-tombench}, MMToM-QA \cite{jin-etal-2024-mmtom}, MOMENTS \cite{villa-cueva-etal-2025-moments}, MuMA-ToM \cite{Shi_Ye_Fang_Jin_Isik_Kuo_Shu_2025}, ToMATO \cite{10.1609/aaai.v39i2.32143}.

\section{Issues Identified in Image Generation} \label{issues_image_generation}
During image generation, we identified several major issues, which are summarized below. Although we manually corrected all identified problems in our dataset, we report them in the appendix to highlight potential pitfalls in generated visual data and to remind future researchers to carefully inspect generated images before use.

\begin{itemize}
\item Incorrect spatial relationships: The model often produces incorrect location information. For example, although the text specifies that the ball is behind the child, the generated image depicts it in an incorrect position.
\item Object and character duplication: The same child may appear multiple times in a single image, or be depicted with extra limbs (e.g., three arms). For instance, when a child moves to a new location, the image may still show the child in the previous location. Similar duplication issues also occur for objects.
\item Unintended text artifacts: Random or meaningless words sometimes appear in the generated images, even when the prompt explicitly specifies that no text should be included.
\item Narrative inconsistencies: Some images conflict with the story context. For example, a mother apologizes for not buying the correct toy, yet the toy is clearly visible and uncovered in the image, allowing the child to see it, which contradicts the narrative.
\item Unnecessary clues: In some cases, the images either explicitly reveal the correct answer or contain misleading elements that may lead to incorrect responses.
\end{itemize}
To ensure the quality of the dataset, we jointly reviewed the text and images in each StoryTurn according to the criteria presented in Table~\ref{validation_criteria}. Table~\ref{image_error_statistics} presents the approximate distribution of the 567 identified image errors. Each image was assigned to the category corresponding to its most prominent error.
\begin{table}[]
\centering
\renewcommand{\arraystretch}{1.1}
\setlength{\tabcolsep}{1.4mm}{
  \resizebox{\columnwidth}{!}{
\begin{tabular}{@{}llcccc@{}}
\toprule
\textbf{Datasets} & \textbf{QA Format} & \textbf{Mental States} & \textbf{Multimodal} & \textbf{Moral} & \textbf{StoryTurn} \\ \midrule
ToMBench          & Multi-Choice       & 6                      & No                  & No             & No                 \\
MOMENTS           & Multi-Choice       & 7                      & Yes                 & No             & No                 \\
MMToM-QA          & Multi-Choice       & 2                      & Yes                 & No             & No                 \\
MuMA-ToM          & Multi-Choice       & 2                      & Yes                 & No             & No                 \\
ToMATO            & Multi-Choice       & 5                      & No                  & No             & No                 \\ \midrule
CoMMET            & Open-Ended         & 7                      & Yes                 & Yes            & Yes                \\ \bottomrule
\end{tabular}}%
}
\caption{\textbf{Comparison with recently published benchmarks that cover multiple mental states defined in ATOMS}. Unlike these benchmarks, we adopt an open-ended format, which is more challenging because no answer options are provided. Additionally, our benchmark includes explicit moral reasoning tasks and evaluates LLMs in conversational settings.}
  \label{compare_with_others}%
\end{table}

\begin{table*}[t]
\centering
\renewcommand{\arraystretch}{1.1}
\setlength{\tabcolsep}{1.5mm}{
  \resizebox{\textwidth}{!}{
    \begin{tabular}{lll}
    \toprule
    \textbf{Criterion}                            & \textbf{Validation Question}                                                                   &  \\
    \midrule
    Absence of unintended text artifacts & Is the image free of unintended text, labels?                                         &  \\
    Correct spatial relationships        & Are all objects and characters located as described in the story?                   &  \\
    Narrative consistency                & Does the image accurately reflect the corresponding StoryTurn?                      &  \\
    Object and character deduplication   & Is each object and character shown only as many times as specified in the story?    &  \\
    Absence of unintended Clues          & Is the image free of unintended visual clues that could reveal the expected answer? &  \\
    \bottomrule
    \end{tabular}}
    }
    \caption{The criteria and validation questions used to evaluate the images in CoMMET.}
      \label{validation_criteria}%
\end{table*}

\begin{table}[t]
\centering
\renewcommand{\arraystretch}{1.05}
\setlength{\tabcolsep}{1.5mm}{
  \resizebox{\columnwidth}{!}{
\begin{tabular}{lcc}
\toprule
\textbf{Revised Error Category}          & \textbf{Count} & \textbf{Percent of 567} \\
\midrule
Unintended text artifacts       & 243   & 42.9\%         \\
Incorrect spatial relationships & 137   & 24.2\%         \\
Narrative inconsistencies       & 77    & 13.6\%         \\
Object/character duplication    & 66    & 11.6\%         \\
Unintended Clues                & 44    & 7.8\%         \\
\bottomrule

\end{tabular}}
}
\caption{The approximate distribution of the 567 identified error images.}
\label{image_error_statistics}
\end{table}

\section{Selection and Validation of the LLM Evaluator} \label{justification_of_llm_evaluator}
To select an appropriate LLM evaluator, we conducted the following validation:
\begin{table}[t]
\centering
\renewcommand{\arraystretch}{1.1}
\setlength{\tabcolsep}{1.5mm}{
  \resizebox{\columnwidth}{!}{
\begin{tabular}{lccc}
\toprule
\textbf{Pair }                               & \textbf{Agree} & \textbf{Disagree} & \textbf{Agreement} \\
\midrule
Gemini   3.1 Pro vs Claude Opus 4.7 & 2496  & 86       & 96.67\%   \\
Gemini   3.1 Pro vs vs GPT 5.4      & 2515  & 67       & 97.41\%   \\
GPT 5.4   vs Claude Opus 4.7        & 2507  & 75       & 97.10\%  \\
\bottomrule
\end{tabular}}
}
\caption{Agreement among LLM judges on Claude Haiku 4.5 outputs for the text-complete subset with and without images.}
\label{llm_agreement_on_claude}
\end{table}

\begin{table}[t]
\centering
\renewcommand{\arraystretch}{1.1}
\setlength{\tabcolsep}{1.5mm}{
  \resizebox{\columnwidth}{!}{
\begin{tabular}{lccc}
\toprule
\textbf{Pair }                      & \textbf{Agree} & \textbf{Disagre}e & \textbf{Agreement} \\
\midrule
Human   vs Gemini 3.1 Pro  & 73    & 16       & 82.02\%   \\
Human vs   Claude Opus 4.7 & 26    & 63       & 29.21\%   \\
Human vs   GPT 5.4         & 49    & 40       & 55.06\%  \\
\bottomrule
\end{tabular}}
}
\caption{Agreement between human experts and LLM judges on Claude Haiku 4.5 responses with inconsistent LLM evaluations in the text-complete subset.}
\label{human_llm_agreement_on_claude}
\end{table}
\begin{itemize}
\item For the text-complete subset, we evaluated the outputs of Claude Haiku 4.5 under two settings: with images and without images. We compared the judgments produced by three strong LLMs: Gemini 3.1 Pro, GPT-5.4, and Claude Opus 4.7. The pairwise agreement among the three judges is reported in Table \ref{llm_agreement_on_claude}. We then manually reviewed 89 "why" questions for which the three judges produced inconsistent evaluations. The agreement between the human annotations and the judgments of each LLM evaluator is reported in Table \ref{human_llm_agreement_on_claude}.

\item For the image-dependent subset, we evaluated the outputs of GPT-4o using the same three LLM judges. Their pairwise agreement is reported in Table \ref{llm_agreement_on_gpt4o}. We also manually reviewed all questions for which the judges produced inconsistent evaluations and report the agreement between each judge and the human annotations in Table \ref{human_llm_agreement_on_gpt4o}.

\end{itemize}
\begin{table}[t]
\centering
\renewcommand{\arraystretch}{1.1}
\setlength{\tabcolsep}{1.5mm}{
  \resizebox{\columnwidth}{!}{
\begin{tabular}{lccc}
\toprule
\textbf{Pair }                               & \textbf{Agree} & \textbf{Disagree} & \textbf{Agreement} \\
\midrule
Gemini   3.1 Pro vs Claude Opus 4.7 & 647   & 38       & 94.45\%   \\
Gemini   3.1 Pro vs vs GPT 5.4      & 646   & 39       & 94.31\%   \\
GPT 5.4   vs Claude Opus 4.7        & 652   & 33       & 95.18\%  \\
\bottomrule
\end{tabular}}
}
\caption{Agreement among LLM judges on GPT-4o outputs for the image-dependent subset.}
\label{llm_agreement_on_gpt4o}
\end{table}

\begin{table}[t]
\centering
\renewcommand{\arraystretch}{1.1}
\setlength{\tabcolsep}{1.5mm}{
  \resizebox{\columnwidth}{!}{
\begin{tabular}{lccc}
\toprule
\textbf{Pair}                       & \textbf{Agree} & \textbf{Disagree} & \textbf{Agreement} \\
\midrule
Human   vs Gemini 3.1 Pro  & 39    & 14       & 73.58\%   \\
Human vs   Claude Opus 4.7 & 24    & 29       & 45.28\%   \\
Human vs   GPT 5.4         & 29    & 24       & 54.72\%  \\
\bottomrule
\end{tabular}}
}
\caption{Agreement between human experts and LLM judges on GPT-4o responses with inconsistent LLM evaluations in the image-dependent subset.}
\label{human_llm_agreement_on_gpt4o}
\end{table}
Two human experts participated in the validation and manual review. The evaluation criteria are relatively straightforward for this task. For example, in Figure \ref{figure_1}, an LLM response is considered correct if it avoids the false inference that the boy knows that his mother did not hear him and instead includes the correct inference that he did not expect to receive the robot.

Overall, the validation results across models from different families, namely Claude Haiku 4.5 and GPT-4o, and across both the text-complete and image-dependent subsets show that Gemini 3.1 Pro achieves the highest agreement with human annotations. These findings support its use as the most accurate and scalable LLM evaluator for our task.
\section{Details of Human Study} \label{human_study}
For this study, we recruited eight volunteers to participate anonymously in the online test without compensation. The study was reviewed by the IRB committee and approved for IRB exemption.

Although the StoryTurns vary across task categories, they are designed to assess the same underlying theory of mind capabilities. Therefore, we selected a representative subset from CoMMET to evaluate adult participants anonymously. For each task category shown in Figure~\ref{data_statistics}, we further considered whether visual information was required, resulting in 20 task--image-requirement combinations. We then constructed eight QA questionnaires. For combinations requiring visual information, we selected one StoryTurn; for combinations not requiring visual information, we selected two StoryTurns. Each questionnaire therefore contained 36 StoryTurns in total. In four questionnaires, images were provided for all StoryTurns, whereas in the other four questionnaires, images were provided only when the textual context was incomplete and visual information was necessary.

We manually examined the human responses and found that adult participants produced more diverse answers than LLMs. In some cases, participants appeared to overthink the questions rather than answering strictly based on the given stories. In other cases, they overlooked important clues, made careless mistakes, or provided overly brief answers that omitted necessary reasoning details. Two detailed examples are provided below for illustration. These observations suggest that the human baseline should be interpreted as a reference for adult performance under our experimental setting, rather than as a strict upper bound on theory of mind ability.
\begin{itemize}
    \item In the belief–desire task shown in Figure 1, which contains five questions, one participant answered the fourth question incorrectly because they overlooked key details, responding: “Not sure, as the mother intended to buy a red race car, but it was sold out. Not sure whether she bought a blue robot or did not buy anything.”
    \item In an intention task, some participants provided overly brief answers to “why” questions. For example, one participant stated only that “you mentioned he wants to show his drawing,” without explicitly explaining that the drawing was in the yellow folder.
\end{itemize}
\section{Performance of Different LLMs on the Original ToM Booklet Tasks}
\label{performance_org}
We evaluated seven LLMs from different model families on the original 39 tasks from the ToM booklet~\cite{Richardson2018DevelopmentOT, SOTOMAYORENRIQUEZ2024109905}. Following the performance analysis on CoMMET, we further divide the tasks according to whether visual information is required. Since the original dataset contains only a small number of StoryTurns for each mental state, we report only the overall performance. Tables~\ref{table 3} and~\ref{table 4} present the performance of different LLMs on the original ToM booklet tasks.

\section{LLMs Perform Well on Some Tasks}
As shown in Tables \ref{text_complete} and \ref{text_incomplete}, LLMs perform well on a subset of tasks, particularly those involving relatively simple mental state reasoning.
\paragraph{Emotion}
This task typically involves a two-turn StoryTurn in which a child’s past experiences or memories are described. LLMs are then asked how the child would feel if they encountered the same scenario again, and what they would do in response. This task is relatively easy for LLMs, and most models achieve strong performance.
\paragraph{Desire}
This task also uses a two-turn StoryTurn to evaluate common and diverse desires. First, LLMs are asked to select one object from a pair that a child prefers. Next, they are told the preference of another child—either the same object or the other one—and asked which object the second child would choose. Almost all LLMs perform perfectly on this task, including the smaller models.

\section{StoryTurn Samples from CoMMET}
\begin{table}[t]
  \centering
  \renewcommand{\arraystretch}{1.1}
\setlength{\tabcolsep}{1.5mm}{
\resizebox{0.57\columnwidth}{!}{
    \begin{tabular}{lc}
    \toprule
    \textbf{Models} & \textbf{Overall} \\
    \midrule
    Llama-32-11B-instruct & 17.19 \\
    LLama-4-Maverick & 18.18 \\
    Claude-Haiku-4.5 & 18.18 \\
    Claude-Opus-4.7 & 45.45 \\
    GPT-4o & 45.45 \\
    GPT-5.4 & 45.45 \\
    Gemini-3.1-Flash-Lite &  63.64\\
    Gemini-3.1-Pro & 72.73 \\
    \bottomrule
    \end{tabular}}}%
    
    \caption{Story-level accuracy of different LLMs on the original ToM booklet task subset where visual informaiton is required to answer the question.}
  \label{table 3}%
\end{table}%

\begin{table}[t]
  \centering
    \renewcommand{\arraystretch}{1.1}
\setlength{\tabcolsep}{1.7mm}{
  \resizebox{\columnwidth}{!}{
    \begin{tabular}{@{}lcc@{}}
\toprule
\textbf{Models}       & \textbf{\begin{tabular}[c]{@{}c@{}}Overall\\ (With Image)\end{tabular}} & \textbf{\begin{tabular}[c]{@{}c@{}}Overall\\ (Without Image)\end{tabular}} \\ \midrule
LLaMA-32-11B-Instruct & 47.12                                                                   & 44.86                                                                      \\
LLaMA-4-Maverick      & 57.14                                                                   & 60.71                                                                      \\
Claude-Haiku-4.5      & 82.14                                                                   & 67.86                                                                      \\
Claude-Opus-4.7       & 71.43                                                                   & 78.57                                                                      \\
GPT-4o                & 75.00                                                                   & 75.00                                                                      \\
GPT-5.4               & 82.14                                                                   & 89.29                                                                      \\
Gemini-3.1-Flash-Lite & 92.86                                                                   & 92.86                                                                      \\
Gemini-3.1-Pro        & 92.86                                                                   & 92.86                                                                      \\ \bottomrule
\end{tabular}
  }
  }
  \caption{Story-level accuracy of different LLMs on the original ToM booklet task subset containing complete textual information.}
  \label{table 4}
\end{table}%

\begin{figure}[h]
    \centering
    \includegraphics[width=\columnwidth]{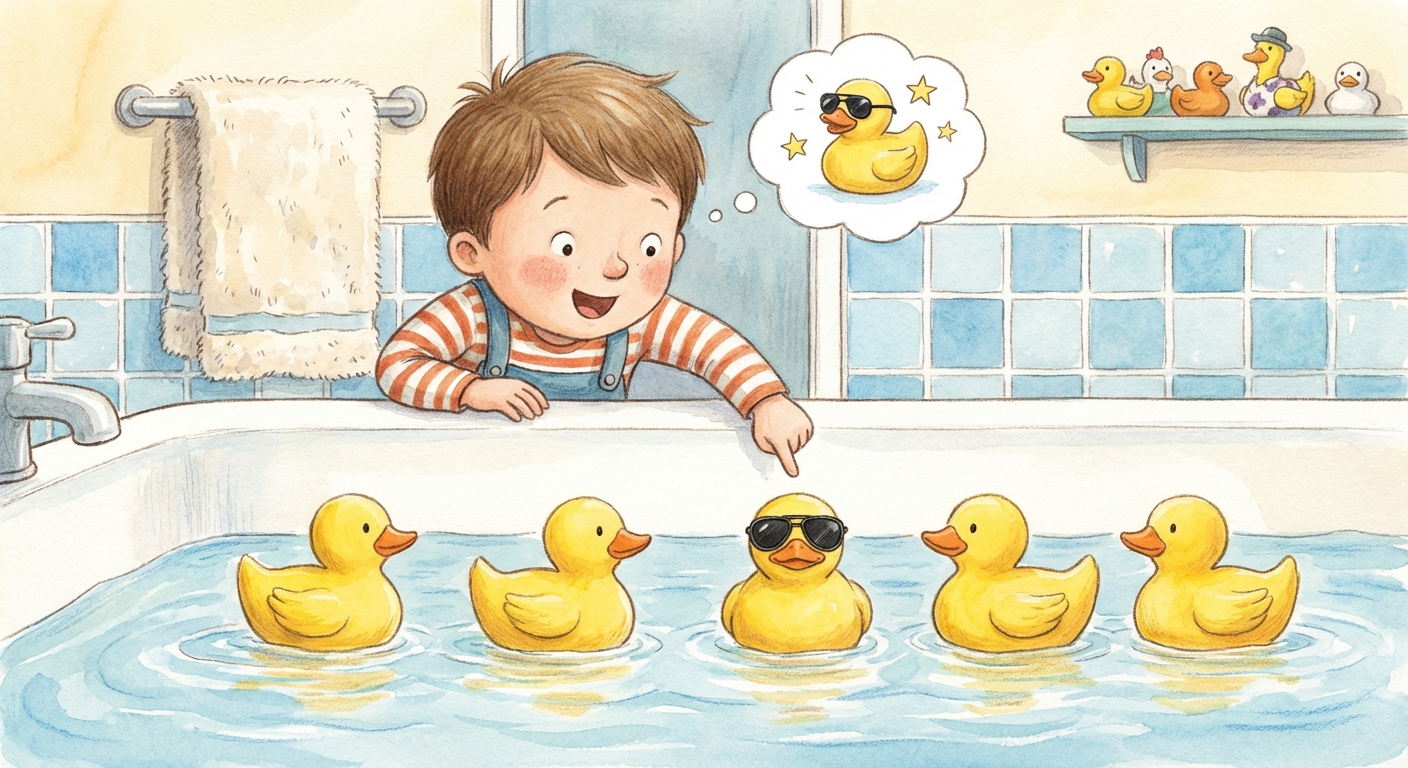}
    \caption{The image for spatial perspective task. ID: story\_5\_2025-12-27 11\_22\_07.738437\_1}
    \label{figure_5}
\end{figure}

\begin{figure}[t]
    \centering
    \includegraphics[width=\columnwidth]{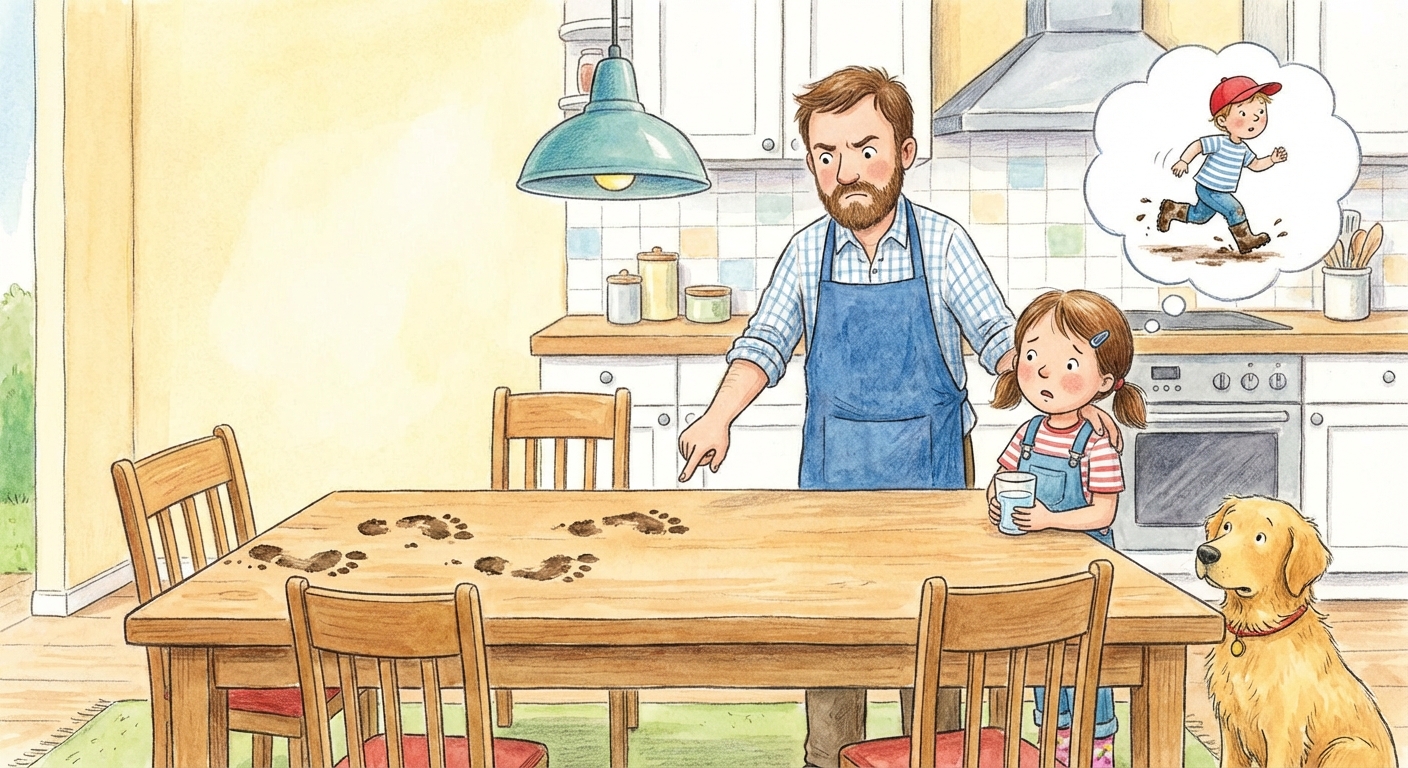}
    \caption{The image for moral reasoning based on false belief task. ID: story\_18\_2025-12-14 09\_16\_30.073127\_img1}
    \label{figure_6}
\end{figure}

\begin{figure}[t]
    \centering
    \includegraphics[width=\columnwidth]{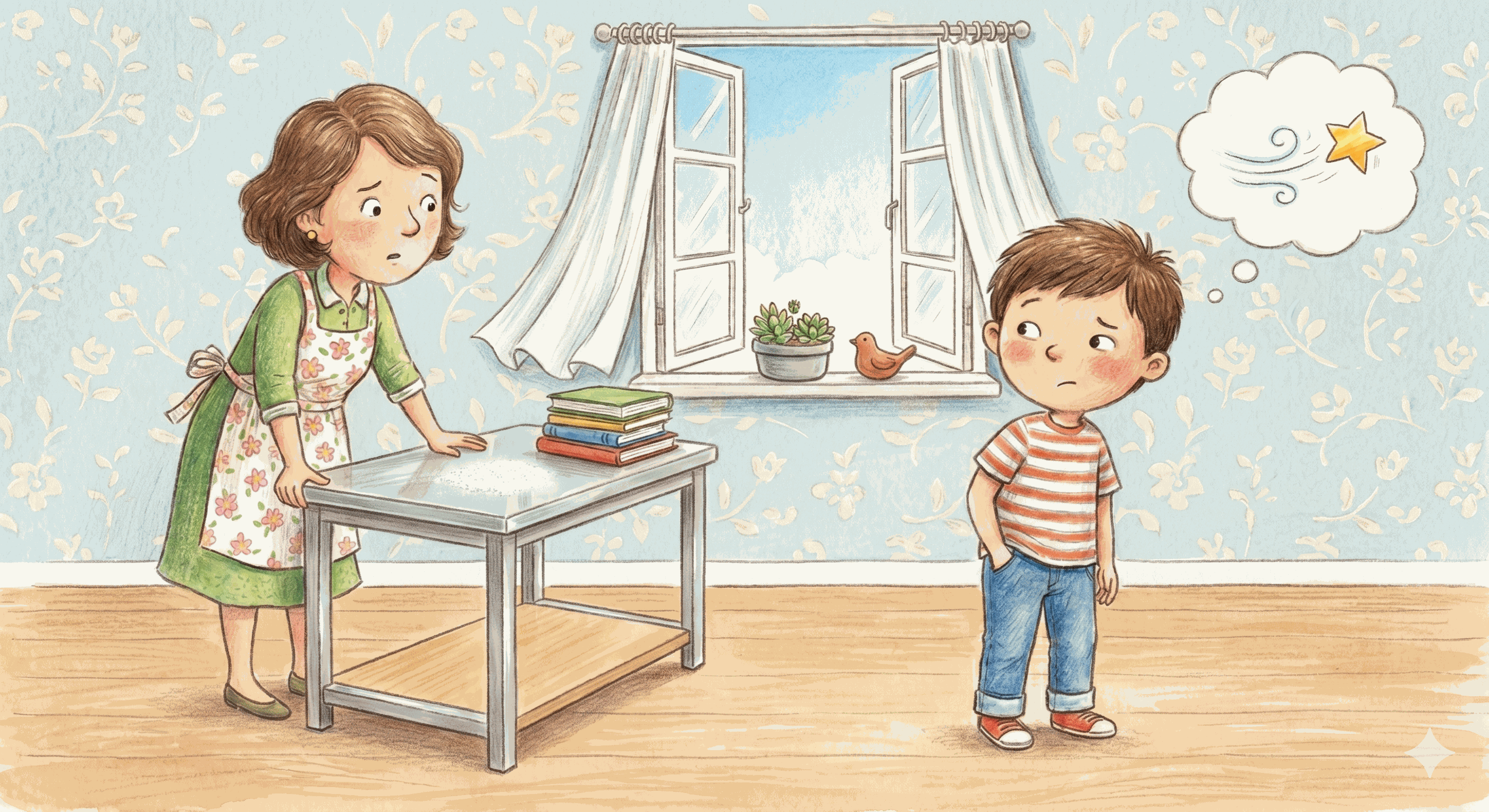}
    \caption{The image for moral reasoning with deception task. ID: story\_36\_2025-12-18 13\_23\_30.289031\_1}
    \label{figure_7}
\end{figure}

Table \ref{table 5} presents selected StoryTurn samples from CoMMET, while Figures \ref{figure_5}, \ref{figure_6}, and \ref{figure_7} provide the corresponding illustrative images for these samples.

\begin{table*}[t]
  \centering
  \footnotesize
  \renewcommand{\arraystretch}{1.15}
  \setlength{\tabcolsep}{6pt}

  \begin{tabularx}{\textwidth}{
    @{}
    >{\raggedright\arraybackslash}p{0.19\textwidth}
    >{\raggedright\arraybackslash}X
    @{}
  }
    \toprule
    \textbf{Task} & \textbf{Sample} \\
    \midrule

    Spatial Perspective &
    "This is Leo. Leo collects rubber ducks. He wants the special duck wearing cool sunglasses. |  | Can you describe its location in the line so I can grab it for him? For example, relative to the left or right, or the middle? | story\_5\_2025-12-27 11:22:07.738437\_1 | The third duck from the left and the third duck from the right, the middle duck." \\

    \addlinespace[3pt]
    \midrule
    \addlinespace[3pt]

    Moral Reasoning based on False Beliefs &
    "It is cleaning day at the house. Dad asks Mia to wipe the dining table. Mia cleans it until it shines. Then, she goes to the kitchen to get a drink of water. While she is gone, her brother Tom runs in with muddy shoes. He climbs on the table, leaves mud everywhere, and runs away. When Mia comes back, Dad walks in too. Dad sees the big muddy footprints and gets mad at Mia for making a mess. |  | Is it fair that Dad is mad at Mia for the dirty table? Why or why not? | story\_18\_2025-12-14 09:16:30.073127\_img1 | No. Because it is not her fault that Tom made the table dirty.",
    \newline
    " |  | Is Dad mad at Tom? Why or why not? |  | No. Because he does not know that Tom climbed on the table with muddy shoes.",
    \newline
    " |  | Is Mia mad at Tom? Why or why not? |  | No. Because she did not see Tom do it, so she does not know he made the mess." \\

    \addlinespace[3pt]
    \midrule
    \addlinespace[3pt]

    Moral Reasoning with Deception &
    "Mrs. Green places a shiny gold sticker on her desk. It is a prize for Mia, who won the math game. Mrs. Green leaves the room for a moment. While she is gone, Leo sneaks up, takes the sticker, and hides it in his pocket. When Mrs. Green comes back, she sees the empty desk. She asks Leo, "Do you know where the sticker went?" Leo tells her, "No, maybe the wind blew it out the window." |  | Is Leo lying? | story\_36\_2025-12-18 13:23:30.289031\_1 | Yes",
    \newline
    " |  | Is the sticker really gone out the window? |  | No",
    \newline
    "Mia comes to the desk to get her prize. She asks Mrs. Green, "Where is my sticker?" Mrs. Green feels bad and says, "I am so sorry Mia, but the wind must have blown it out the window. It is gone." |  | Is Mrs. Green lying? Why or why not? |  | No. Because she does not know Leo has it; she thinks the wind took it",
    \newline
    " |  | Is the sticker actually lost outside? |  | No",
    \newline
    " |  | How does Mia feel, Sad or Fine? |  | Sad",
    \newline
    " |  | Even though the sticker is safe in Leo's pocket, Mia feels sad. Why? |  | Because she didn't know; she thought the sticker was lost forever",
    \newline
    "Leo feels guilty, so he puts the sticker back on the desk. Just then, a big gust of wind comes and actually blows the sticker out the open window! Now the sticker is really gone. |  | Was Leo lying earlier when he told Mrs. Green the wind blew the sticker away? Why or why not? |  | Yes. Because at the time he told Mrs. Green the wind took it, the sticker was actually in his pocket" \\

    \bottomrule
  \end{tabularx}

  \caption{StoryTurn Samples from CoMMET}
  \label{table 5}%
\end{table*}%


\section{Additional Tasks Challenging for MLLMs: Non-Literal Communication and Emotion}
Taking the sarcasm and non-sarcasm tasks as examples, each StoryTurn contains two questions. The first asks why the child says a particular utterance (e.g., "Mmm, my favorite") in either unhappy or happy scenarios. The second question asks about the child’s emotional state. We find that LLMs—including GPT-5.4—answer some sarcasm questions incorrectly. Smaller models perform substantially worse; for example, LLaMA-3.2-11B-Vision-Instruct achieves only 4.55\% accuracy on the sarcasm task.

\section{More Experiment on Feedback} \label{experiment_on_feedback}

\begin{table}[]
\centering
\renewcommand{\arraystretch}{1.1}
\setlength{\tabcolsep}{1.5mm}{
  \resizebox{\columnwidth}{!}{
\begin{tabular}{@{}lccl@{}}
\toprule
\textbf{Model}          & \textbf{\begin{tabular}[c]{@{}c@{}}Text-Complete\\ Subset (with Images)\end{tabular}} & \textbf{\begin{tabular}[c]{@{}c@{}}Image-Dependent\\ Subset\end{tabular}} & \textbf{Feedback} \\ \midrule
\multirow{2}{*}{GPT-4o} & 6/7                                                                                   & 17/37                                                                     & Original          \\
                        & 6/8                                                                                   & 23/36                                                                     & Correctness-Only  \\ \bottomrule
\end{tabular}}%
}
\caption{Effectiveness of Original and Correctness-Only Feedback in Correcting LLM Response Trajectories. We report the proportion of cases in which an LLM answered a question incorrectly but answered the subsequent question correctly after receiving feedback, among all cases where an incorrect response was followed by feedback.}
\label{comparison_of_feedbacks}
\end{table}

We added a correctness-only control and compared it with the original feedback condition wherever applicable, as shown in Table \ref{comparison_of_feedbacks}. Following the setting in Table \ref{feedback}, we report the proportion of cases in which an LLM answered a question incorrectly but then answered the subsequent question correctly after receiving feedback, among all cases where an incorrect answer was followed by feedback. We observed that, in some cases, correctness-only feedback encouraged the model to focus more closely on the story context. For example:
\begin{itemize}
    \item Original feedback: "Oh look, Dad is actually buying the striped hat."
    \item GPT-4o's response to the subsequent question after receiving the original feedback: "Maybe Dad liked the striped hat better or thought it was a better choice."
    \item Correctness-only feedback: "Your answer is wrong."
    \item GPT-4o's response to the subsequent question after receiving the correctness-only feedback: "Because the striped hat was next to the mirror."
    \item Expected answer: "Because it is next to the mirror."
\end{itemize}
\end{document}